\documentclass[twoside]{article} \usepackage[accepted]{aistats2017}

\usepackage{graphicx}
\usepackage{subfigure} 
\usepackage{amsmath}
\usepackage{algorithm}
\usepackage[round]{natbib}
\usepackage{algorithmic}
\usepackage{amsfonts}
\usepackage{booktabs}
\usepackage{todonotes}
\usepackage{microtype}
\usepackage[capitalize]{cleveref}
\usepackage{mathtools}
\usepackage{microtype}
\DeclarePairedDelimiter{\diagfences}{(}{)}
\newcommand{\diag}{\operatorname{diag}\diagfences}

\begin{document}

%

%

\twocolumn[

\aistatstitle{Recurrent Collective Classification}

\aistatsauthor{ Shuangfei Fan \And Bert Huang  }

\aistatsaddress{ Dept.~of Electrical and Computer Engineering\\Virginia Tech \And Dept.~of Computer Science\\Virginia Tech  } ]

\begin{abstract}
We propose a new method for training iterative collective classifiers for labeling nodes in network data. The iterative classification algorithm (ICA) is a canonical method for incorporating relational information into classification. Yet, existing methods for training ICA models rely on the assumption that relational features reflect the true labels of the nodes. This unrealistic assumption introduces a bias that is inconsistent with the actual prediction algorithm. In this paper, we introduce recurrent collective classification (RCC), a variant of ICA analogous to recurrent neural network prediction. RCC accommodates any differentiable local classifier and relational feature functions. We provide gradient-based strategies for optimizing over model parameters to more directly minimize the loss function. In our experiments, this direct loss minimization translates to improved accuracy and robustness on real network data. We demonstrate the robustness of RCC in settings where local classification is very noisy, settings that are particularly challenging for ICA.
\end{abstract}

\section{Introduction}
\label{sec:intro}

Data science tasks often require reasoning about networks of connected entities, such as social and information networks. In classification tasks, the connections among network nodes can have important effects on node-labeling patterns, so models that perform classification in networks should consider network structure to fully represent the underlying phenomena. For example, when classifying individuals by their personality traits in a social network, a common pattern is that individuals will communicate with like-mined individuals, suggesting that predicted labels should also tend to be uniform among connected nodes. 
Collective classification methods aim to make predictions based on this insight. In this paper, we introduce a collective classification framework that will enable an algorithm to more directly optimize the performance of trained collective classifiers.

Iterative classification is a framework that enables a variety of supervised learning methods to incorporate information from networks. The base machine learning method can be any standard classifier that labels examples based on input features. The iterative classification algorithm (ICA) operates by using previous predictions about neighboring nodes as inputs to the current predictor. This pipeline creates a feedback loop that allows models to pass information through the network and capture the effect of structure on classification. In spite of the feedback loop being the most important aspect of ICA, existing approaches train models in a manner that ignores the feedback-loop structure. In this paper, we introduce \emph{recurrent collective classification} (RCC), which corrects this discrepancy between the learning and prediction algorithms, incorporating principles used in deep learning and recurrent neural networks into the training process.

Existing learning algorithms for iterative classification resort to an approximation based on the unrealistic assumption that the predicted labels of neighbors are their true classes \citep{neville:srl00,london:book13}. This assumption is overly optimistic.  If it were true, iteration would be unnecessary. Because the assumption is overly optimistic, it causes the learned models to cascade and amplify errors when the assumption is broken in early stages of prediction. In contrast, ICA uses predicted neighbor labels as feedback for each subsequent prediction, which means that if the model was trained expecting these predicted labels to be perfect, it will not be robust to situations where predictions are noisy or inaccurate. In this paper, we correct this faulty assumption and develop an approach that trains models for iterative classification by treating the intermediate predictions as latent variables. We compute gradients to the classification loss function using back-propagation through iterative classification.

To compute gradients for ICA, we break down the ICA process into differentiable (or sub-differentiable) operations. In many cases, the base classifier is differentiable with respect to its parameters. For example, if it is a logistic regression, it has a well-studied gradient. ICA also computes dynamic relational features using the predictions of network neighbors. These relational features are also functions through which gradients can be propagated. Finally, because the same base-classifier parameters should be used at all iterations of ICA, we can use methods for recurrent neural networks such as back-propagation through time (BPTT) \citep{werbos:ieee90} to compute the combined gradient. In contrast with existing strategies for training ICA, the resulting training optimization more closely mimics the actual procedure that ICA uses for prediction.

The RCC framework accommodates a variety of base classifiers and relational feature functions. The only restriction is that they must be differentiable. Therefore, RCC is nearly as general as ICA, and its prediction procedure is practically identical to ICA. The key difference is that the view of the algorithm as nested, differentiable functions enables a training procedure that is better aligned with RCC and ICA prediction. 

We evaluate RCC on data where collective classification has previously been shown to be helpful. We demonstrate that RCC trains classifiers that are robust to situations where local predictions are inaccurate.

\section{Related Work}
\label{sec:related}

Node classification is one of the fundamental tasks in analysis of network data \citep{getoor:exploration05,lindamood2009inferring}. \emph{Collective classification} addresses this task by making joint classifications of connected nodes \citep{kong2011multi,taskar2002discriminative,neville2003collective}. Gibbs sampling (GS) is another approach for collective classification using the iterative classification framework \citep{mcdowell2007cautious,sen:aimag08}, that introduces randomization into the iterative classification. ICA and GS have been shown repeatedly to be effective frameworks for collective classification \citep{sen:aimag08,jensen:sigkdd04,neville:srl00,lu:icml03,mcdowell2007cautious}. 
One of the more natural motivations for collective classification comes from the study of social networks, where the phenomenon of \emph{homophily}---the tendency of individuals to interact with other similar individuals---has been an important concept \citep{mcpherson:ars01,bilgic2007combining}. The types of dependencies that can exist in networks are not limited to assortative relationships, and methods such as ICA enable models to encode both assortative and non-assortative phenomena.

Collective classification has been studied in both inductive and transductive settings. In inductive settings, a collective classifier is typically trained on a fully labeled training network and evaluated at test time on a completely new network with no known labels. In transductive settings, the classifier is both trained and tested on partially labeled networks, or possibly in the same network with a different set of labels known during each phase \citep{xiang:aistats11}. In this paper, we focus on the inductive setting.

Through the interpretation of non-terminal classifications as latent variables, ICA can be related to deep learning methods. Since the same classifier is used to predict each latent layer, ICA is most related to recurrent neural networks (RNNs), which feature a similar feedback loop in which an output of a neural network is used as its input in a subsequent iteration. RNNs were introduced decades ago \citep{angeline1994evolutionary,connor1994recurrent}, but they have recently become prominent because of their effectiveness at modeling sequences, such as those occurring in natural language processing, e.g., \citep{socher:icml11,mikolov2010recurrent,graves2014towards,graves2012supervised}. A now standard method for gradient optimization of RNN parameters is known as back-propagation through time \citep{werbos:ieee90,hochreiter2001gradient,ishikawa1996structural}, which unrolls recurrent networks and computes gradients for the parameters separately before combining them into a single update. 

After the first version of our manuscript was published \citep{fantraining}, two other groups have independently and concurrently pursued very similar directions, applying deep learning to collective classification \citep{pham2016column,moore2017deep}. \citet{moore2017deep} proposed to use RNNs for node-based relational classification tasks. They transform each node and its set of neighbors into an unordered sequence and use an LSTM-based RNN to predict the class label as the output of that sequence. \citet{pham2016column} proposed another deep learning model for collective classification. Their approach is fully end-to-end, with a neural network that computes hidden units connected in the same structure as the input network, but the hidden units do not output class probabilities. Instead, they are abstract representations of learned local and relational features. These concurrent studies represent different ideas for bringing the power of neural networks to collective classification.

\section{Iterative Classification}
\label{sec:ica}

\begin{algorithm}[tb]
   \caption{The Iterative Classification Algorithm}
   \label{alg:example}
\begin{algorithmic}[1]
   \STATE {\bfseries Input:} Adj.~matrix ${\bf A}$, node features ${\bf X}$, num.~of iterations $T$, classifier $f$, and relational feature function $g$.
   \STATE Initialize labels $Y$ 
   \COMMENT {e.g., uniform probability}
   \FOR{$t$ from 1 to $T$}
   \STATE ${\bf R} \leftarrow g({Y}; {\bf A})$ 
   \COMMENT{Compute relational features}.
   \STATE ${Y} \leftarrow f({\bf X}, \bf{R}; {\bf \Theta})$ \COMMENT{Compute new predictions}.
   \ENDFOR
\STATE \textbf{return} labels ${Y}$
\end{algorithmic}
\end{algorithm}

In this section, we review the \emph{iterative classification algorithm} (ICA) and the standard method for training its parameters. ICA provides a framework for node classification in networks. ICA is given a decorated graph $G = \{V, E, {\bf X}\}$, where $V = \{v_1, \ldots, v_n\}$, $E$ contains pairs of linked nodes $(v_i, v_j) \in E$, and each node is associated with a respective feature vector ${\bf x}_i \in {\mathbb R}^d \equiv {\cal X}$, where ${\bf X} = \{{\bf x}_1, \ldots, {\bf x}_n\}$. Using these inputs, ICA outputs a set of predictions $ Y $ classifying each of the nodes in $V$ into a discrete label space $\cal Y$. Throughout this paper, we will consider the multi-class setting, in which the labels can take one of $k$ class-label values. ICA makes the label predictions by iteratively classifying nodes by their \emph{local features} ${\bf x}_i$ and their dynamic \emph{relational features} ${\bf r}_i$, which is in a common space ${\bf r}_i \in {\cal R}$. In other words, the classifier is a function $f$ that maps ${\cal X} \times {\cal R}$ to $\cal Y$, parameterized by a parameter variable ${\bf \Theta}$.

ICA first initializes labels as $Y^{(0)}$, and it then iterates the steps

\begin{equation}
\begin{aligned}
{\bf R}^{(t-1)} &\leftarrow g(Y^{(t-1)}; {\bf A}) \\
Y^{(t)} &\leftarrow f({\bf X}, {\bf R}^{(t-1)}; {\bf \Theta})
\end{aligned}
\label{eq:iter}
\end{equation}
from iterations $t = 1$ to $t = T$. 

The dynamic relational features $\bf R$ where $ {\bf R} = \{{\bf r}_1, \ldots, {\bf r}_n\}$ are computed based on the current estimated labels of each node. They enable the classifier to reason about patterns of labels among connected nodes. E.g., a common relational feature is the average prediction of neighboring nodes. 

Using any such aggregation statistic creates a relational feature vector of dimensionality $k$, where each entry is the occurrence rate of its corresponding label in the node's neighbors. I.e.,, the relational features are computed by a feature function $g$ that maps ${\cal Y}^n$ to ${\cal R}^n$. 

Since the dynamic relational features are computed based on the output of the classifier, the entire process is iterated: (1) all nodes are labeled by the classifier $f$ using the current dynamic relational features, then (2) the dynamic relational features are computed with $g$ based on the new labels. These two phases are repeated either until the predictions converge and do not change between iterations or until a cutoff point. 

The ICA framework is general in that any classifier $f$ can be used and any form of a dynamic relational feature function $g$ can be used. In practice, researchers have used naive Bayes, logistic regression, support vector machines as classifiers, and they have used averages, sums, and presence as relational features \citep{sen:aimag08,jensen:sigkdd04,neville:srl00,lu:icml03}. The generality of the framework is a key benefit of ICA---one that we aim to preserve in our proposed framework. In contrast to specially designed methods for collective classification, the modularity of ICA makes it a flexible meta-algorithm.

In many settings, we consider real-valued local and relational features, so each node is described by a feature vector created by concatenating its local features with its relational features $[{\bf x}_i, ~ {\bf r}_i]$. In this case, it is convenient to notate the classification function in matrix form. Let $\bf X$ denote the feature matrix for the graph, such that the $i$th row of ${\bf X}$, i.e., ${\bf x}_i$ is the (transposed) feature vector for node $v_i$. Similarly, let $\bf R$ denote the relational feature matrix, such that the $i$th row of $\bf R$ is the dynamic relational feature vector of node $v_i$. For convenience, we consider the case where one type of dynamic relational feature is used, meaning the dimensionality of $\bf R$ is $n$ by $k$ (though it is easy to extend both ICA and RCC to multiple relational features).

The training procedure for ICA trains the classifiers by computing relational features using the training labels. Given a training set consisting of a set of nodes $V$, edges $E$, node features $X$, and ground-truth labels $\hat Y$, one generates relational features $R$ using the true training labels, creating fully instantiated, fully labeled inputs for the classifier. The model parameters $\bf \Theta$ are learned by setting $\hat{{\bf R}} = g({\hat  Y})$, then using $\bf X$ and $\hat{{\bf R}}$ as the input to a supervised training scheme appropriate for fitting the parameters ${\bf \Theta}$ of $f$.

This training methodology is only correct in the situation where we expect a \emph{perfect-classification fixed-point}. In such a fixed point, the relational features are computed using the true labels, and the classifiers perform perfectly, exactly predicting the true labels of the nodes; the relational features are computed, using the perfectly predicted labels, to be exactly the same features that are computed using the true training labels; since the relational features are computed using the exactly predicted true labels, the output is the same as in the first step, and the algorithm converges perfectly to the true labels. Using the matrix form of the relational feature computation above, this absurd fixed point would be characterized as
${Y} = f \left( {\bf X},~ g({\hat  Y}); {\bf \Theta}\right)$.

Unfortunately, such a fixed point is unrealistic. In practice, the classifications can be inaccurate or made with low confidence due to a lack of reliable local information. Thus, training the model to expect that the neighbor labels be perfect creates an overconfidence that can lead to cascading errors. 

\section{Recurrent Collective Classification}
\label{sec:rcc}

We propose the recurrent collective classification (RCC) framework to address the previously mentioned deficiencies in ICA training. The RCC framework computes derivatives for the stages of collective classification and provides a general scheme for how to compute the gradient of the loss function with respect to the classifier parameters. \Cref{alg:rcc} summarizes RCC gradient computation. This gradient computation enables the training of collective classifiers in a manner that more directly mimics how they will be applied. At test time, a collective classifier is often given only local features of nodes connected in a network. Thus, any relational features are derived from \emph{predicted} neighbor labels. A classifier considering these neighbor labels should therefore consider common patterns of misclassification of neighbor labels. In contrast to the ICA training procedure, which invites the classifier to become overly reliant on the verity of the neighbor labels.

\begin{algorithm}[tb]
   \caption{RCC Gradient Computation}
   \label{alg:rcc}
\begin{algorithmic}[1]
   \STATE {\bfseries Input:} Graph $G = \{V, E\}$, number of iterations $T$, classifier $f$, and relational feature function $g$.
\STATE Initialize parameter ${\bf \Theta}$ set ${\bf P}^{(0)} = 0$
\FOR{$t$ from 1 to $T$} 
   \STATE ${\bf R}^{(t)} \leftarrow g({\bf P}^{(t-1)}; {\bf A})$ 
   \COMMENT{Relational features}.
   \STATE ${\bf P}^{(t)} \leftarrow f({\bf X}, \bf{R}^{(t)}; {\bf \Theta})$ \COMMENT{New predictions}
   \ENDFOR
\STATE $\bf \Delta^{(T)} \leftarrow L'({\bf P}^{(T)})$
\COMMENT{Loss gradient for output}
\FOR{$t$ from $T$ to $2$} 
\FOR{$j$ from $1$ to $N$}
\STATE ${\bf \delta}_j^{(t-1)} \leftarrow  \sum_{i: (i,j) \in E} {\bf \delta}_i^{(t)} \cdot f'\left( {\bf r}_i^{(t-1)} \right) \cdot g'_i \left( {\bf p}_j^{(t)}\right)$
\ENDFOR
\ENDFOR
\STATE $\nabla({\bf \Theta}) \leftarrow \sum_{t = 1}^T {\bf \Delta}^{(t)} f'({\bf \Theta})$
\end{algorithmic}
\end{algorithm}

\begin{figure}[tb]
\vskip 0.2in
\begin{center}
\centerline{\includegraphics[width=0.4\textwidth]{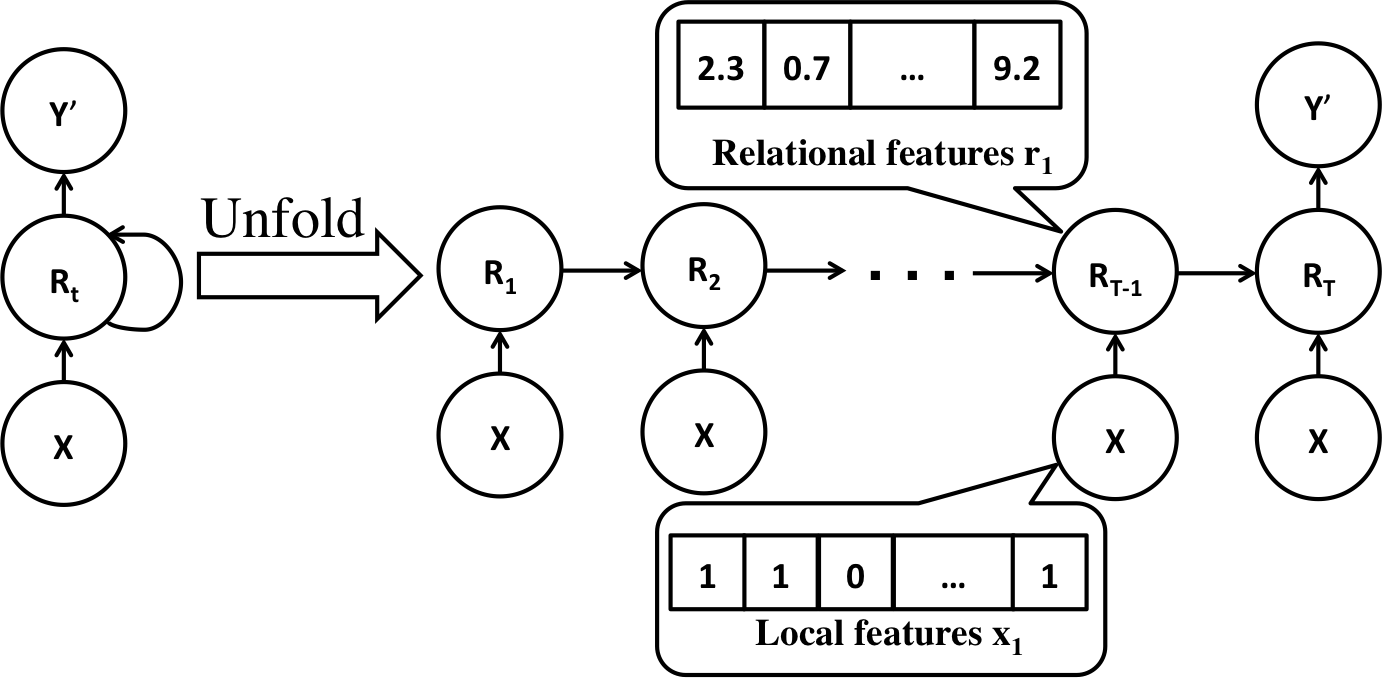}}
\caption{Structure of RCC/ICA prediction. The recurrent form (left) unrolls into a form (right) that explicitly considers each iteration as a separate operation.}
\label{fig:rcc}
\end{center}
\vskip -0.2in
\end{figure}

Prediction in RCC is analogous to ICA. We initialize ${\bf P}^{(0)}$ (e.g., setting it to all zeros, or to random values), then iterate the steps
${\bf R}^{(t-1)} \leftarrow g({\bf P}^{(t-1)}; {\bf A})$, and
${\bf P}^{(t)} \leftarrow f({\bf X}, {\bf R}^{(t-1)}; {\bf \Theta})$
from iterations $t = 1$ to $t = T$. This recursive procedure can be interpreted as a recurrent neural network, as illustrated in \Cref{fig:rcc}.

RCC training requires that the local classifier function $f$ is equipped with efficiently computable gradients with respect to the parameters ${\bf \Theta}$ and the relational features ${\bf R}$, and that the relational feature function $g$ is equipped with an efficiently computable gradient with respect to the current predictions ${\bf P}$. Since these functions map matrices to matrices, their gradients are computed via matrix calculus. However, as we discuss in \Cref{sec:structure}, they often have sparse structure that makes their computation efficient. 
As a shorthand, with some abuse of notation, we refer to these gradients using the following definitions:
\begin{equation}
\begin{aligned}
f'_{(t)}({\bf \Theta}) &:= \partial f({\bf X}, {\bf R}^{(t-1)}; {\bf \Theta}) / \partial {\bf \Theta}\\
f'({\bf R}^{(t-1)}) &:= \partial f({\bf X}, {\bf R}^{(t-1)}; {\bf \Theta}) / \partial {\bf R}^{(t-1)}\\
g'({\bf P}^{(t)}) &:= \partial g({\bf P}^{(t)}; {\bf A}) / \partial {\bf P}^{(t)}.
\end{aligned}
\end{equation} 
At iteration $T$, we evaluate a loss function $L$ on the final prediction ${\bf P}^{(T)}$. RCC admits any differentiable loss function. For example, for multinomial distribution, we can use Softmax classifier, which uses the cross-entropy loss. The cross-entropy loss is to minimize the cross-entropy between the estimated distribution ${\bf q}$ and the "true" distribution ${\bf p}$ as the loss function. It is defined as:

\begin{equation}
\begin{aligned}
L = H(p,q) = -\sum_{x}p(x)\log q(x)
\end{aligned}
\label{eq:cross_entropy}
\end{equation}

Denote the gradient of the loss with respect to the predictions ${\bf P}^{(t)}$, i.e., ${\bf \Delta}^{(t)} := \frac{\partial L}{\partial {\bf P}^{(t)} }$.
Given the gradients for each iteration, the gradient of the loss with respect to the parameters ${\bf \Theta}$ is $\frac{\partial L}{\partial {\bf \Theta}} = \sum_{t = 1}^T {\bf \Delta}^{(t)} f'({\bf \Theta}).$

Based on these formulas, RCC is able to use any gradient-based optimization over the parameter space, so long as it is able to compute the loss gradients ${\bf \Delta}$. These gradients can be computed using chain rule and dynamic programming, as in back-propagation. The general formula is:

\begin{equation}
\begin{aligned}
{\bf \Delta}^{(t-1)} &= \frac{\partial L}{\partial {\bf P}^{(t-1)}} = \left( \frac{\partial L}{\partial {\bf P}^{(t)}} \right) \left( \frac{\partial {\bf P}^{(t)}}{\partial {\bf P}^{(t-1)}} \right) \\
&= 
{\bf \Delta}^{(t)} 
\left( \frac{\partial {\bf P}^{(t)}}{\partial {\bf R}^{(t-1)}} \right) 
\left( \frac{\partial {\bf R}^{(t-1)}}{\partial {\bf P}^{(t-1)}} \right).
\end{aligned}
\label{eq:fullmatrix}
\end{equation}

Though these matrix gradients can be large for arbitrary matrix functions, collective classification has a particular structure that enables efficient computation.

\subsection{Derivative Structure}
\label{sec:structure}

Key aspects in understanding the efficient computability of these gradients are the independence relationships among the input and output variables. These independence relationships are universal to any local classifier and to any relational feature.

Because the local classifier operates on each node independently, in each iteration, there is no dependence between the classification of any node and the relational features of any other node. The matrix derivative of the classifier function $f$ is block diagonal, only having nonzero derivative between each node's relational-feature row ${\bf r}_i$ and its output-prediction row ${\bf p}_i$.

Because relational features are completely determined by the neighbors of each particular node, there is only dependence between the predictions of any node $j$ and the relational feature of node $i$ if they are neighbors in the graph. The matrix derivative of the relational feature function $g$ therefore has a block structure in the shape of the sparse adjacency matrix. 

Considering both of these sparse block structures, we can define a new shorthand for the nonzero elements of the matrix derivatives:

\begin{equation}
\begin{aligned}
f' \left( {\bf r}_i^{(t-1)} \right) &:= \partial f({\bf x}, {\bf r}_i^{(t-1)}; {\bf \Theta}) / \partial {\bf r}_i^{(t-1)} 
\hspace{-0.01in} \equiv \hspace{-0.01in}
\partial {\bf p}_i^{(t)} / \partial {\bf r}_i^{(t-1)} \\
g'_i \left( {\bf p}_j^{(t)} \right) &:= \partial \left[g({\bf p}_j^{(t)}; {\bf A})\right]_i / \partial {\bf p}_j^{(t)} \equiv \partial {\bf r}_i^{(t)} / \partial {\bf p}_j^{(t)}.
\end{aligned}
\end{equation}
\label{eq:derivative}

\Cref{fig:gradients} illustrates the sparse structure of the full matrix gradients using these definitions.
The matrix chain rule then simplifies to

\begin{equation}
\begin{aligned}
{\boldsymbol \delta}_j^{(t-1)} &= \sum_{i: (i,j) \in E} \left( \frac{\partial { L}}{\partial {\bf p}_i^t} \right) \left( \frac{\partial {\bf p}_i^t}{\partial {\bf r}_i^{(t-1)}} \right) \left( \frac{\partial{\bf r}_i^{(t-1)}}{\partial {\bf p}_j^{(t-1)}} \right)\\
{\boldsymbol \delta}_j^{(t-1)}&= \sum_{i: (i,j) \in E} {\boldsymbol \delta}_i^{(t)} \cdot f'\left( {\bf r}_i^{(t-1)} \right) \cdot g'_i \left( {\bf p}_j^{(t)}\right).\\
\end{aligned}
\end{equation}
\label{eq:delta}

where the loss derivatives ${\boldsymbol \delta}_i^{(t)}$ are gradient vectors and the classifier and feature function derivatives, $f'$ and $g'$ are small Jacobian matrices. This structure is significantly sparser and more efficient to compute than the full matrix calculus in \Cref{eq:fullmatrix}.

\begin{figure*}[tbp]
{
\tiny
\begin{minipage}{.5\textwidth}
\begin{align*}
\frac{\partial {\bf P}^{(t)}}{\partial {\bf R}^{(t-1)}} = \left[ 
\begin{array}{cccc}
\left[ f'({\bf r}_1^{(t-1)}) \right] & \left[ {\bf 0} \right] & \hdots & \left[ {\bf 0} \right] \\
\left[ {\bf 0} \right] & \left[ f'({\bf r}_2^{(t-1)}) \right] & \hdots & \left[ {\bf 0} \right]  \\
\vdots & \vdots  & \ddots & \vdots \\
\left[ {\bf 0} \right] & \left[ {\bf 0} \right] & \hdots & \left[ f'({\bf r}_N^{(t-1)})  \right]  \\
\end{array}
\right]
\end{align*}
\end{minipage}
\begin{minipage}{.5\textwidth}
\begin{align*}
\frac{\partial {\bf R}^{(t-1)}}{\partial {\bf P}^{(t-1)}} = \left[ 
\begin{array}{cccc}
\left[ {\bf 0} \right] & \left[ g'_1({\bf p}_2^{(t-1)}) \right] & \left[ {\bf 0} \right] & \hdots \\ 
\left[ g'_2({\bf p}_1^{(t-1)}) \right] & \left[ {\bf 0} \right] &  \left[ g'_2({\bf p}_3^{(t-1)}) \right]  & \hdots \\
\left[ {\bf 0} \right] & \left[ g'_3({\bf p}_2^{(t-1)}) \right] & \left[ {\bf 0} \right] & \hdots \\
\vdots & \vdots & \vdots & \ddots \\
\end{array}
\right]
\end{align*}
\end{minipage}
}
\caption{Matrix gradients for RCC. The left gradient is block-diagonal, since predictions depend only on each node's relational features. The right gradient has block sparsity matching the sparsity of adjacency matrix ${\bf A}$.}
\label{fig:gradients}
\end{figure*}

\subsection{Example Local Classifiers}

One benefit of ICA is that it can use any local classifier and relational feature. Similarly, given \Cref{eq:delta}, we can also easily use many local classifiers $f$ and relational features $g$ into RCC. In this section, we give some examples of these configurations.

First, we consider linear classifiers with activation functions where the linear product prediction scores are squashed by different activation functions. A common activation function is the \textbf{logistic sigmoid function}:
\begin{equation}
f(\bf{x_i}, {\bf r_i}^{(t-1)}; {\bf \Theta}) = \tfrac{1}{1 + \exp\left(-\left[ \bf{x}_i, \quad \bf{r}_i^{(t-1)} \right]\cdot {\bf \Theta}\right)} = {\bf p_i}.
\end{equation}
where we write $[ {\bf x}_i, {\bf r}_i]$ to indicate the horizontal concatenation of the row vectors ${\bf x}_i$ and ${\bf r}_i$.
Let ${\bf \Theta} = \left[\begin{array}{c} {\bf \Theta}_{\bf x} \\ {\bf \Theta}_{\bf r} \end{array} \right]$, separating the parameters for the relational features to submatrix ${\bf \Theta}_{\bf r}$. The derivative with respect to ${\bf r}_i^{(t-1)}$ is the Jacobian matrix
$f'({\bf r_i}^{(t-1)}) = \diag{{\bf p}_i^{(t-1)} (1 - {\bf p}_i^{(t)})} {\bf \Theta}_{\bf r}^\top$.
The gradient with respect to the parameters is
\begin{equation}
f'({\bf \Theta}) = - \diag{{\bf p}_i^{(t-1)} (1 - {\bf p}_i^{(t)})} \cdot \left[{\bf X}, {\bf R}^{(t-1)}\right] ~.
\end{equation}

Another activation function is the \textbf{tempered softmax function}, i.e., a generalization of both the normalized multi-class logistic and the max function:
\begin{align}
f(\bf{x_i}, {\bf r_i}^{(t-1)}; {\bf \Theta}) &= \tfrac{\exp\left(\left(\left[ \bf{x_i}, \quad \bf{r}_i^{(t-1)} \right]\cdot {\bf \Theta}\right)/{\tau} \right)}{\mathbf{1}^\top \exp\left(\left(\left[ \bf{x_i}, \quad \bf{r}_i^{(t-1)} \right]\cdot {\bf \Theta}\right)/{\tau} \right)} = {\bf p_i}.
\label{equ:softmax}
\end{align}
where $\tau$ is the temperature parameter. As $\tau$ approaches 0, the limit of the softmax is an argmax indicator vector indicating the maximal entry, and at $\tau = 1$, this is exactly the multi-class logistic class probability.

The Jacobian of the softmax with respect to the relational features is
$f'({\bf r_i}^{(t-1)}) = \frac{1}{\tau} (\diag{{\bf p_i}} - {\bf p_i}{\bf p_i}^\top) {\bf \Theta}_{\bf r}^\top.$ And the gradient with respect to the parameters is
\begin{equation}
f'({\bf \Theta}) = - \frac{1}{\tau} (\diag{{\bf p_i}} - {\bf p_i}{\bf p_i}^\top) \cdot \left[{\bf X}, {\bf R}^{(t-1)}\right]~.
\end{equation}

The local classifier used in RCC can also seamlessly incorporate non-linear classifiers such as multi-layer perceptrons and neural networks that have well-understood derivatives.

\subsection{Example Relational Features}
There are many aggregation operators can be used to define the relational features. Commonly used features include the {\bf sum} of probabilities for each class, the {\bf proportion} of each class in the neighborhood,  {\bf mode}, which is the class label with the highest probability among neighbors and {\bf exists}, which is an indicator for each class label \citep{sen:aimag08}. The choice of which one to use depends on the application. Here we discuss three of them and provide their derivatives.

The sum aggregation function $g$ can be written as ${\bf r_i} = g({\bf P}^{(t)}; {\bf a}_i) =  {\bf a}_i^{\top}{\bf P}^{(t)}$,
where ${\bf a_i}$ is the adjacency vector of node $i$. The Jacobian of the sum feature is $g'_i({\bf p}^{(t)}_{j}) =  a_{ij} {\bf I}_k$,
where ${\bf I}_k$ is the identity matrix.

The proportion operator is similar to sum, except that we scale the adjacency matrix ${\bf A}$ by normalizing each row vector. The operation can be summarized as 
$\hat {\bf A} = {\bf A} \oslash ({\bf A} \cdot {\bf 1}_{N})$, where ${\bf 1}_{N}$ is the all-ones matrix of size $N$ and $N$ is the total number of nodes. The operator $\oslash$ performs element-wise division.
The proportion operator uses $\hat {\bf A} $ for the relational feature function
${\bf r_i} = g({\bf P}^{(t)}; \hat {\bf a}_i) =  \hat {\bf a}_i^{\top}{\bf P}^{(t)}$,
where $\hat {\bf a_i}$ is the normalized adjacency vector of node $i$. The Jacobian for the proportion feature function is $g'_i({\bf p}^{(t)}_{j}) =  \hat {a}_{ij} {\bf I}_k$. 

For the mode aggregation operator, we can use the tempered softmax function, providing a differentiable form
${\bf r_i} = g({\bf P}^{(t)}; {\bf a}_i) = \tfrac{\exp({\bf a}_i^{\top}{\bf P}^{(t)}/{\tau} )}{\mathbf{1}^\top \exp({\bf a}_i^{\top}{\bf P}^{(t)}/{\tau} )}
$ where small $\tau$ values closely mimic the true mode. The Jacobian of the mode aggregation relational feature is
$g'_i({\bf p}^{(t)}_{j}) =  (1 / \tau) a_{ij}\cdot (\diag{{\bf r}_i} - {\bf r}_i {\bf r}_i^\top)$.

\subsection{Computational Complexity}

Prediction in RCC (and ICA) requires computing relational features and predicting $T$ times. For most relational features, each node must perform $O(k)$ work per neighbor, amounting to $O(|E| k)$ total computation. Assuming a constant number of relational features, the linear local classifier performs $O(d + k)$ work. Thus each prediction iteration requires $O(|E| k + d)$ time, and the full prediction requires $O(T(|E| k + d))$ time. 

Gradient-based optimization of the RCC objective function is a non-convex program, so the total number of iterations for learning is nontrivial to analyze in general. However, we can analyze the computational complexity of each gradient computation. The computation of the gradient requires three phases: the computation of the classifier and relational feature derivatives, the computation of the loss derivatives (back-propagation), and the computation of the parameter derivatives.

The cost of computing the classifier and relational feature derivatives depends on the chosen classifier and relational features. In the examples provided earlier, the cost is $O(k^2)$ for both.

The back-propagation formula in \Cref{eq:delta} computes a single row of the ${\bf \Delta}^{(t)}$ derivative, which has $N$ total rows. The inner summation iterates over the neighbors of the current node, performing a vector-matrix product and a matrix-matrix product. The derivatives are Jacobians of size $O(k)$ by $O(k)$, so the total cost is $O(k^2)$. Thus, the combined cost is $O(T |E| k^2)$ for back-propagating the derivative through $T$ iterations.

Finally, computation of the gradient for ${\bf \Theta}$ also depends on the local classifier. For the linear classifiers described earlier, the cost is $O(dk)$ per iteration, so $O(Tdk)$. Therefore, the entire gradient computation costs $O(T(dk + |E|k^2))$, which is comparable to the cost of prediction itself, since $k$ is often a small quantity.

\section{Experiments}
\label{sec:exp}

In this section, we describe experiments that test whether the proposed training method for RCC is able to improve upon existing methods of training iterative classifiers. We explore scenarios where the local classifier produces inaccurate predictions, challenging the faulty assumption implied by training ICA and Gibbs sampling (GS) with relational features computed using the true labels. The results illustrate that our hypothesis is correct, identifying a variety of settings where RCC better optimizes the training objective and produces more accurate predictions on held-out data.

\begin{figure}[tb]
\vskip 0.2in
\begin{center}
\includegraphics[width=3in, trim={.5in .5in .5in .3in}, clip]{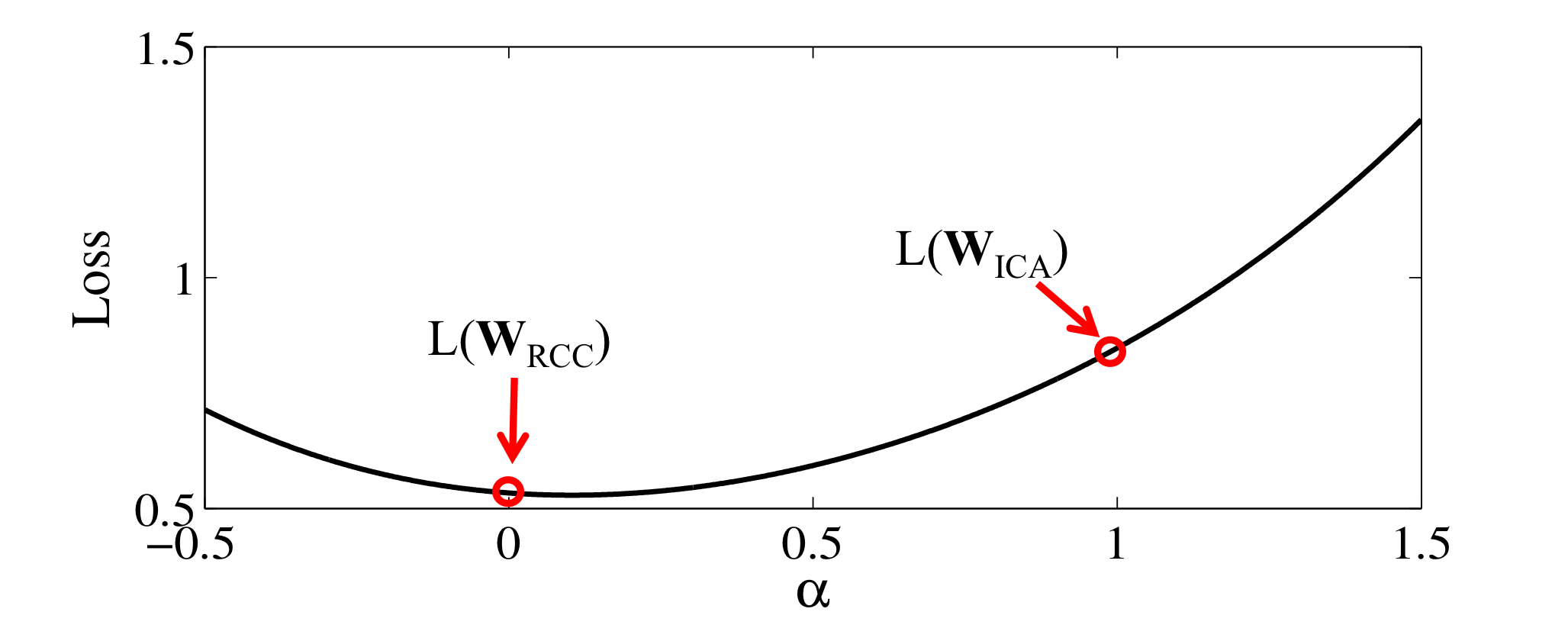}
\caption{Cross-section of the training objective. Circles are solutions from RCC and ICA training. The RCC solution is at a local minimum while ICA's is not.}
\label{fig:optimacompare}
\end{center}
\vskip -0.2in
\end{figure}

\begin{figure*}[htbp]
\centering
\begin{minipage}[]{0.24\textwidth}
\begin{center}
\tiny Cora\\
\includegraphics[width=1\textwidth]{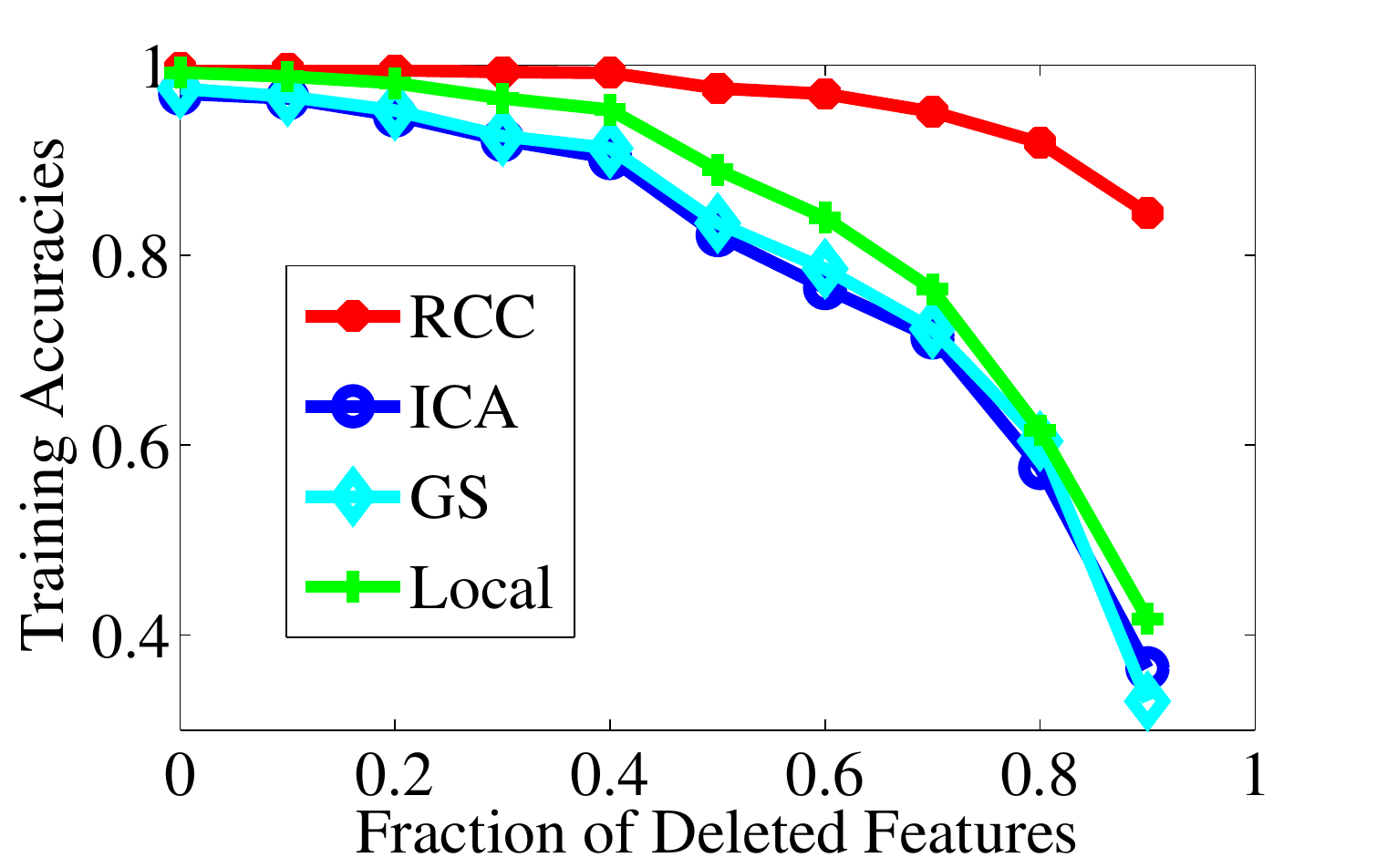}\\
\includegraphics[width=1\textwidth]{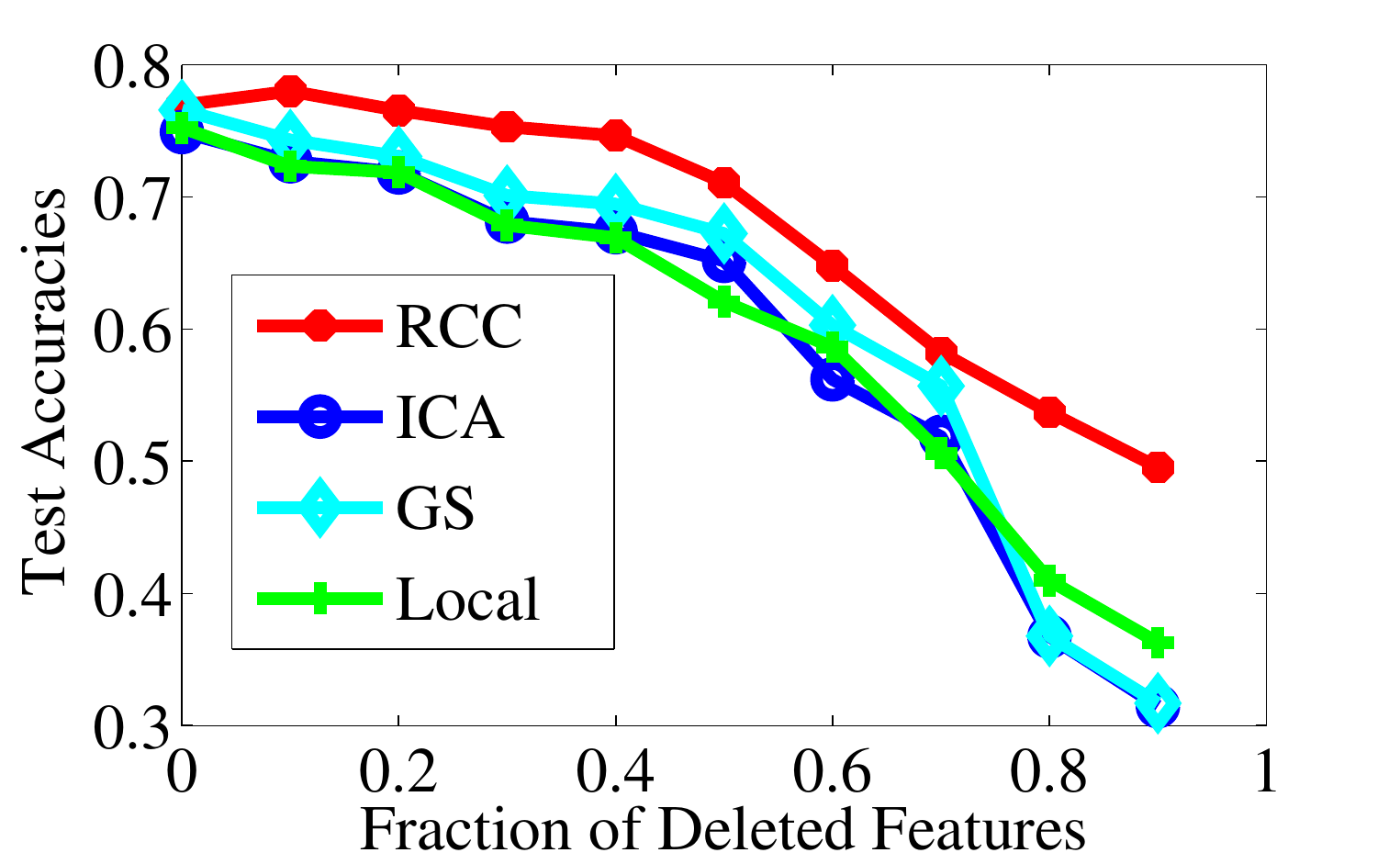} 
\end{center}
\end{minipage}
\begin{minipage}[]{0.24\textwidth}
\begin{center}
\tiny Citeseer\\
\includegraphics[width=1\textwidth]{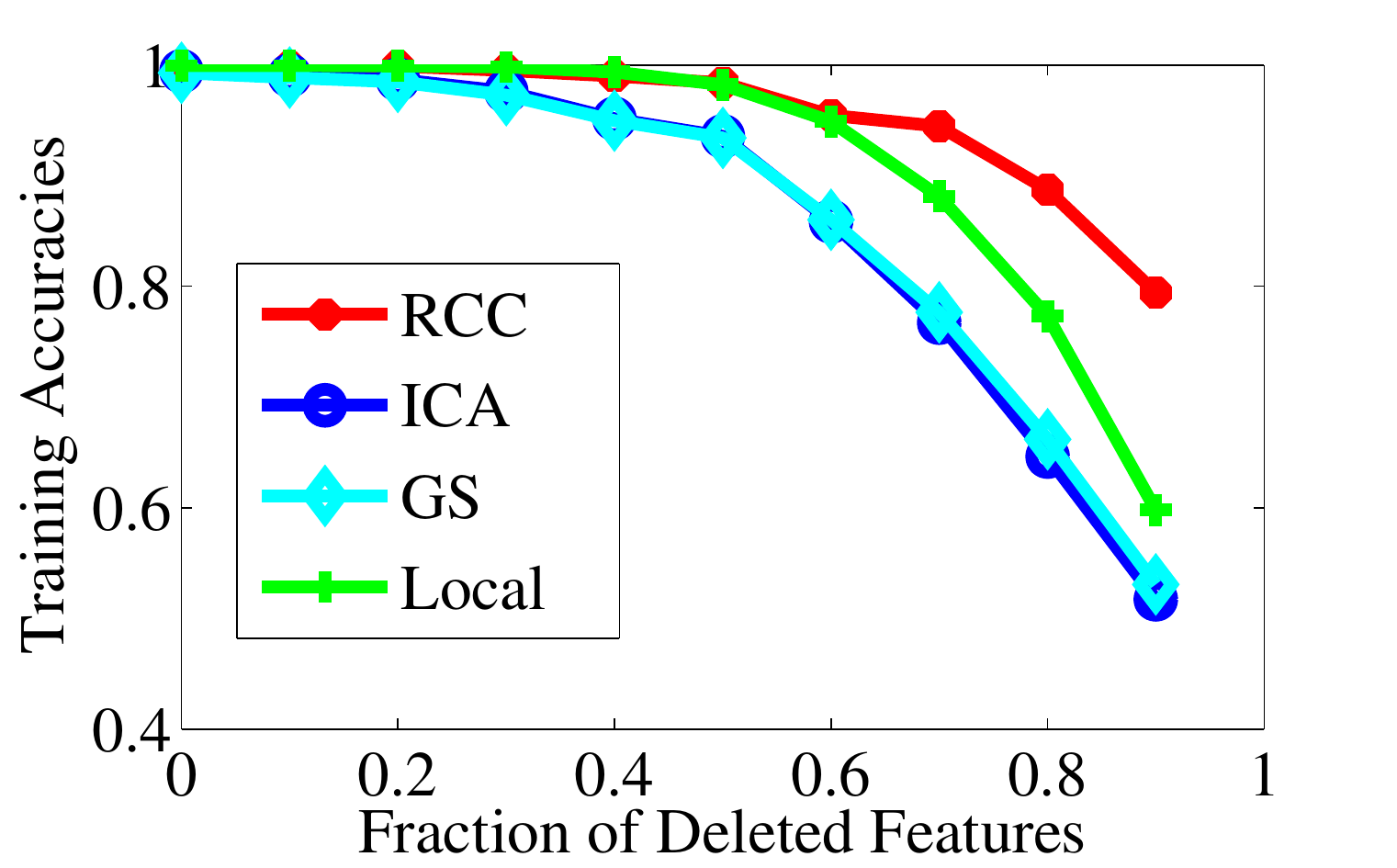}\\
\includegraphics[width=1\textwidth]{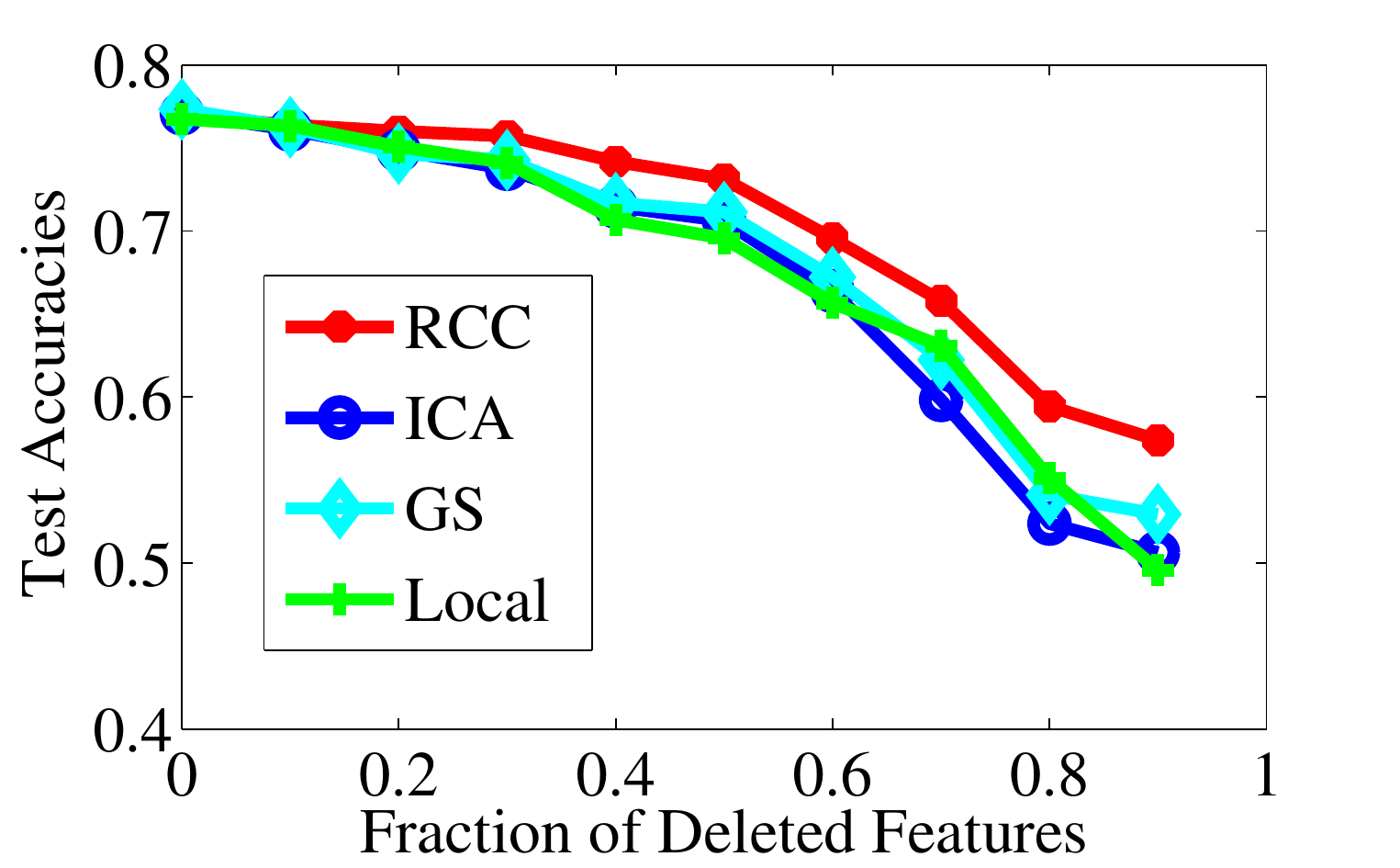} 
\end{center}
\end{minipage}
\begin{minipage}[]{0.24\textwidth}
\begin{center}
\tiny Facebook\\
\includegraphics[width=1\textwidth]{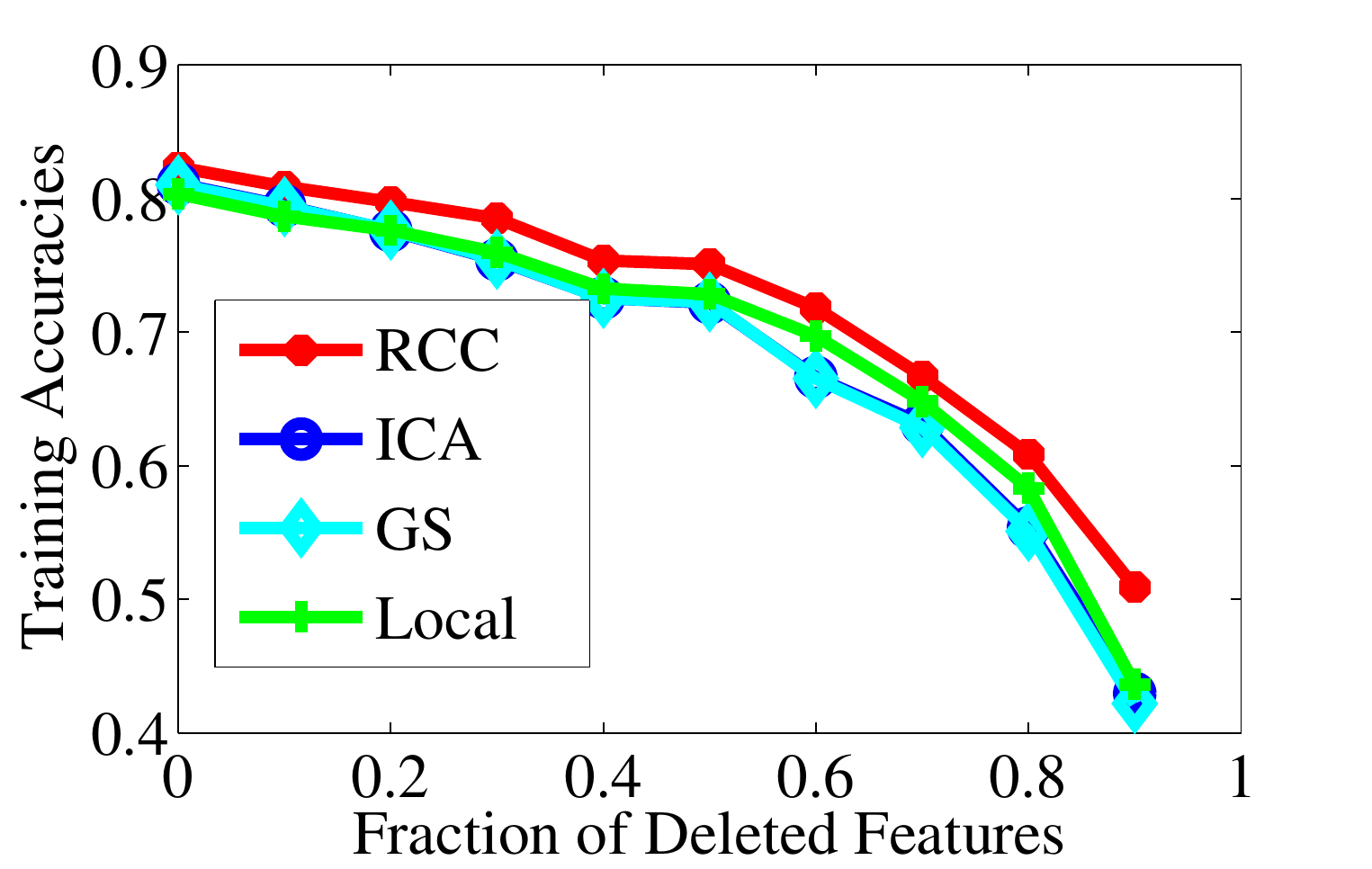}\\
\includegraphics[width=1\textwidth]{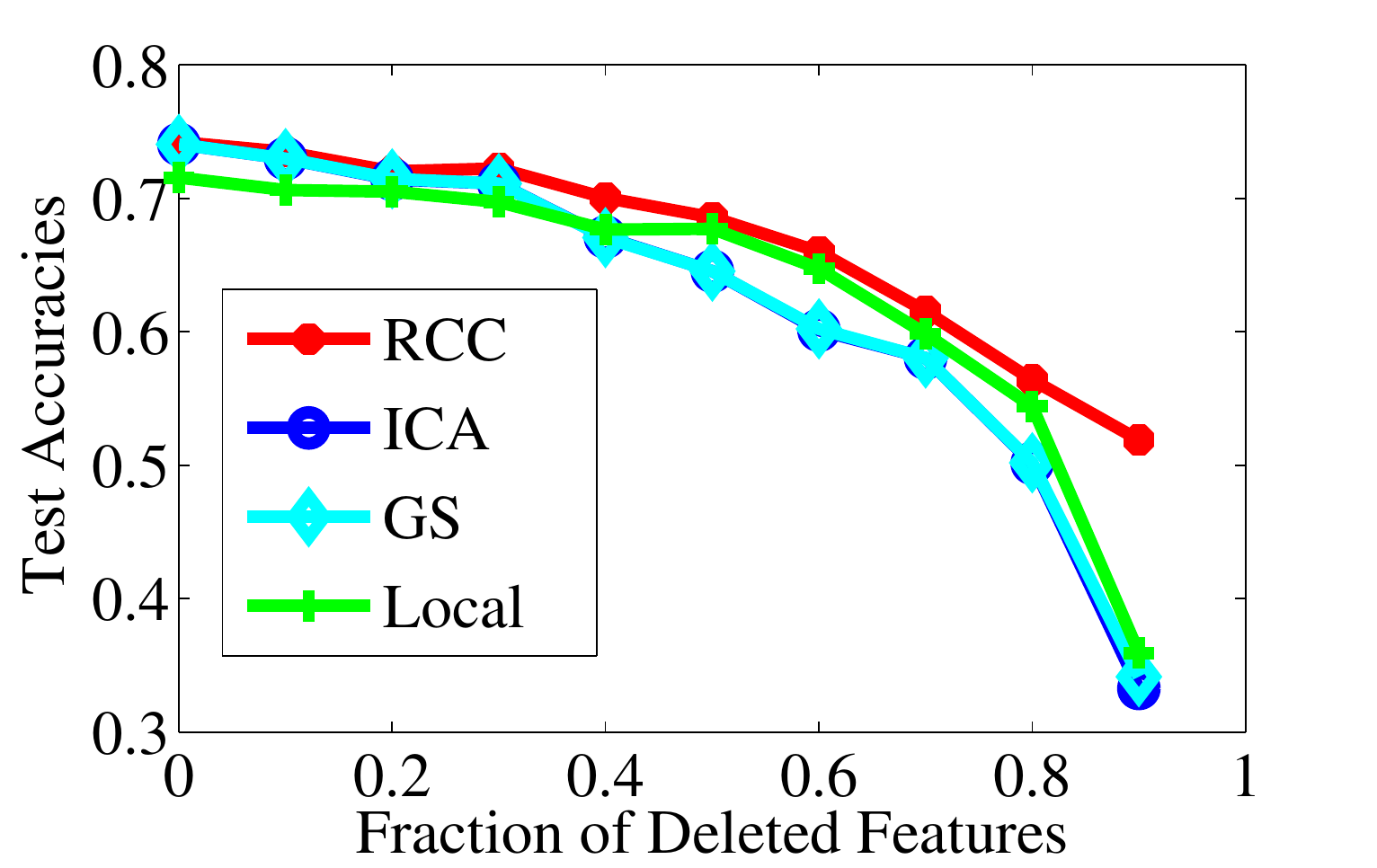} 
\end{center}
\end{minipage}
\begin{minipage}[]{0.24\textwidth}
\begin{center}
\tiny Weizmann\\
\includegraphics[width=1\textwidth]{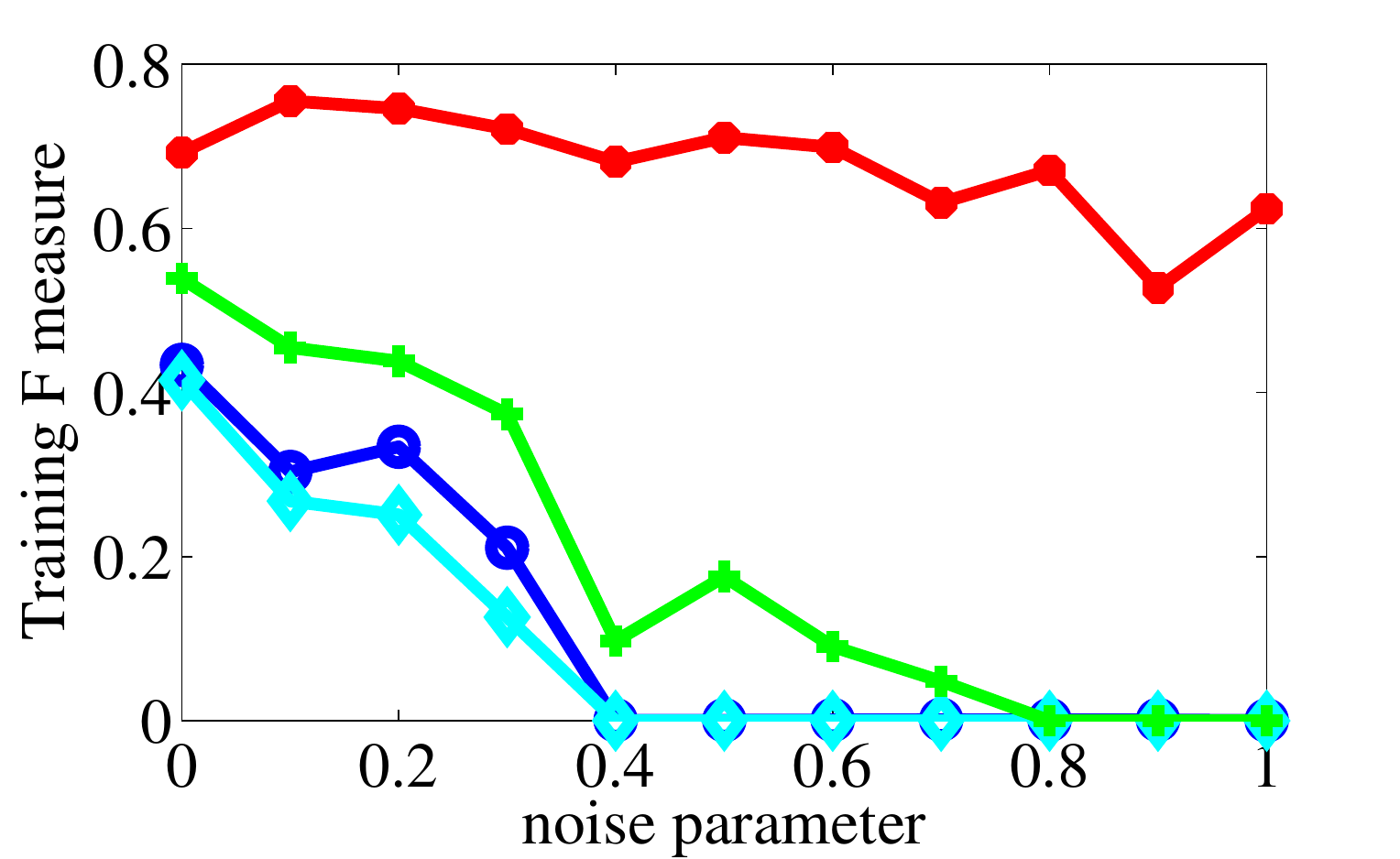}\\
\includegraphics[width=1\textwidth]{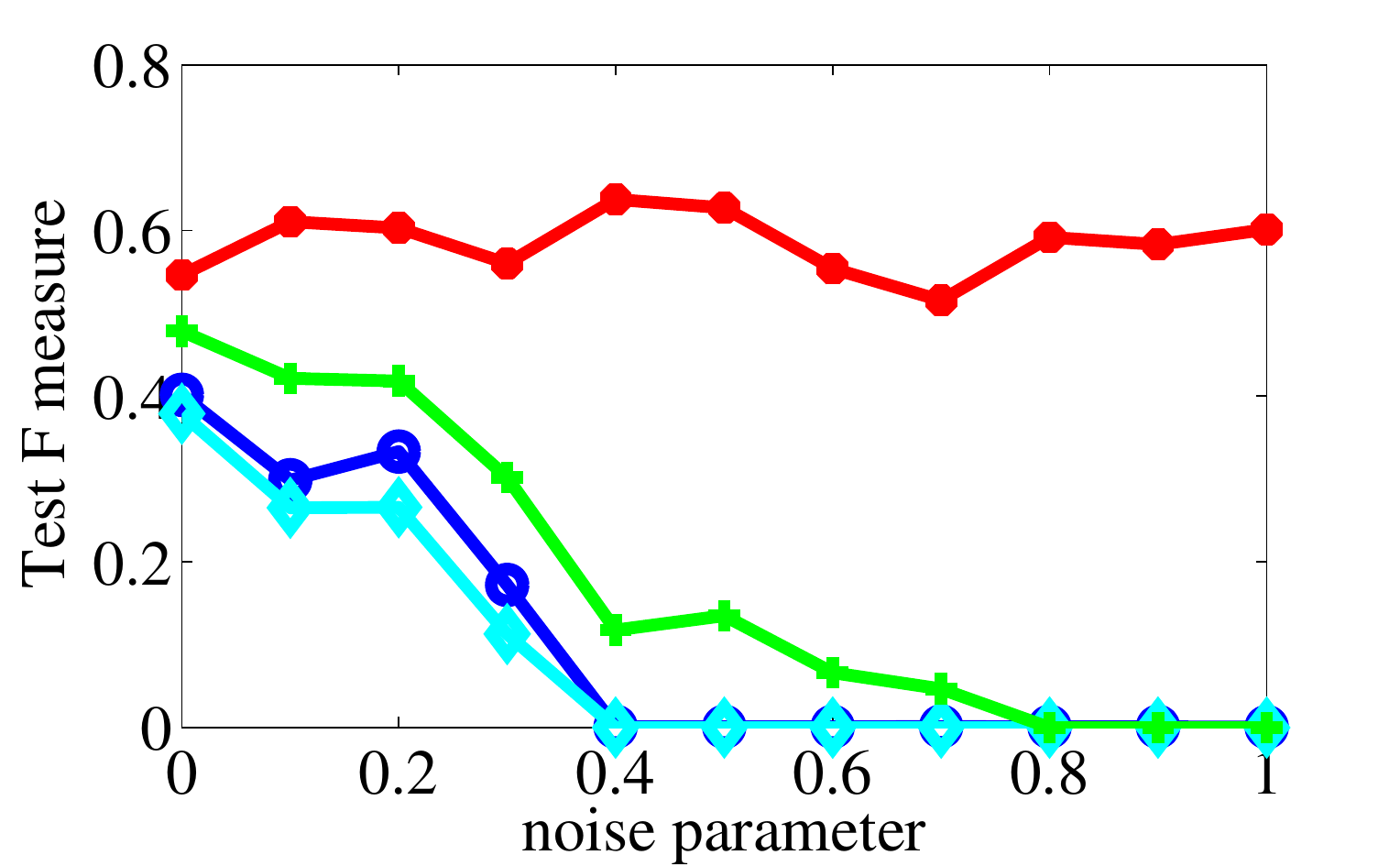} 
\end{center}
\end{minipage}
\caption{Performance of collective classifiers on the four tasks. Each curve plots the average training or testing accuracy or F-measure over the amount of noise (feature removal or salt-and-pepper). RCC dominates all methods on training accuracy, and it performs significantly better in testing than others when there is weak local signal.}
\label{fig:acc}
\end{figure*}

\paragraph{Data}

We experimented with four data sets, two bibliographic data sets, one social network data set and one image dataset. The Cora data set is a collection of 2,708 machine learning publications categorized into seven classes \citep{sen:aimag08}. Each publication in the dataset is described by a 0/1-valued word vector indicating the absence/presence of the corresponding word from the dictionary. The dictionary consists of 3,703 unique words.
The CiteSeer data set is a collection of 3,312 research publications crawled from the CiteSeer repository \citep{sen:aimag08}. It consists of 3,312 scientific publications categorized into six classes. Each publication in the data set is described by a 0/1-valued word vector indicating the absence/presence of the corresponding word from the dictionary. 
The dictionary consists of 1,433 unique words.
The social network data we use is the Facebook ego network data \citep{mcauley:nips12}, which includes users' personal information, friend lists, and ego networks. We combine the ego networks and node features to form a single connected network, with 4,039 users' anonymized profile data and links between them. We used the feature ``education type,'' as a label, aiming to predict how many degrees each user has, i.e., the four classes are whether their highest level of education is secondary, undergraduate, or graduate, or unknown.
The image dataset we use is the Weizmann horse-image segmentation set \citep{borenstein2004learning}. We subsample 20 images of horses on various backgrounds. We use features described by \citet{domke2013learning} for each pixel: We expand the RGB values of each pixel and the normalized vertical and horizontal positions into 64 features using sinusoidal expansion. The goal is to identify which pixels are part of a horse and which are background. We pose this task as a collective classification by constructing a grid graph over neighboring pixels.
Specifically, we expand the feature vectors by multiplying our basic feature vector with all binary vectors $\bf c$ of the appropriate length and take the $sin({\bf c} \times s)$ and $cos({\bf c} \times s)$ as our final feature vectors, which have 64 features in total.

\paragraph{Setup}

For each experiment, we evaluate on four different approaches for node classification: (1) local prediction using only the local features; (2) ICA trained using the true labels; (3) GS trained using the true labels; and (4) RCC trained using back-propagation.They are all trained using a multi-class logistic regression loss function. ICA and GS are trained with the concatenated relational features computed from the true labels in addition to local features as input. RCC is trained using the training labels only in computing the loss, but never as input to the classifier in any form.

For each of the learning objectives, we optimize using the adagrad approach \citep{duchi:jmlr11}, in which gradients are rescaled based on the magnitude of previously seen gradients. For the gradient ${\bf g}_\tau$ at optimization iteration $\tau$, one updates the variable ${\bf \Theta}$ with
${\bf \Theta}_\tau \leftarrow {\bf \Theta}_{\tau - 1} - \eta  \frac{{\bf g}_\tau}{\sqrt{\sum_{i = 1}^{\tau} {\bf g}_{i} \odot {\bf g}_{i}}}$,
where the gradient division by the historical magnitude is elementwise. Adagrad is one of many approaches that has been shown in practice to accelerate convergence of gradient-based optimization. Training is done with 2,000 iterations of adagrad and an initial learning rate of $\eta = 0.1$. We evaluate performance of each method using a range of regularization parameter settings from $1 \times 10^{-3}$ to $1$. 

For the document and social network data, we perform snowball sampling to extract a random $1/5$ of the nodes to hold out as a isolated test network. We train on the induced graph of the $4/5$ remaining nodes, and measure predictive accuracy on both the training graph and testing graph. For the image data, we train on two random splits of 10 training and 10 testing images. The training accuracy should more closely reflect whether each method's training strategy effectively optimizes the model to fit the observed data, and the testing accuracy should additionally reflect how well the learned model generalizes. We compute both training and testing accuracy by feeding the learned model only the local features and link structure, meaning that though ICA and GS is typically \emph{trained} with the training labels as input, we do not provide the labels to them when \emph{evaluating} its training accuracy. Our hypothesis is that by directly computing the gradient of the actual prediction procedure for collective classification, RCC will produce better training performance, which should translate to better testing performance.

\paragraph{Empirical Evaluation of Loss}

To illustrate that RCC training optimizes the training loss function better than training with the true labels, we can compare the training loss associated with the soft-max output loss. We first train model weights using (1) RCC and (2) the ICA approach using the true labels. We then define a tradeoff function 
$l(\alpha) = L({\bf \Theta}_{\text{RCC}} + \alpha({\bf \Theta}_{\text{ICA}}  - {\bf \Theta}_{\text{RCC}} ))$,
where $L$ is the loss function, ${\bf \Theta}_{\text{RCC}}$ is the parameter matrix from the model trained by RCC, ${\bf \Theta}_{\text{ICA}}$ is the weight matrix trained by the ICA strategy. We apply the loss function to different weight matrices on the line that connects ${\bf \Theta}_{\text{RCC}}$ and ${\bf \Theta}_{\text{ICA}}$. This curve is a cross-section of the high-dimensional function that is the actual training loss. When $\alpha = 0$, the value is the loss of the RCC weights $L({\bf \Theta}_{\text{RCC}})$, and (2) when $\alpha=1$, the value is that of the ICA weights $L( {\bf \Theta}_{\text{ICA}})$. The results for Cora are shown in \Cref{fig:optimacompare}. For all four data sets, the loss obtained by RCC training $L({\bf \Theta}_{\text{RCC}})$ is near a local minimum and the loss obtained by ICA training is not. Unsurprisingly, these results suggest that the training procedure of RCC is a better strategy to optimize the loss function than using the true-label relational features, corroborating our proposed approach. 

\begin{table}[!t]
\renewcommand{\arraystretch}{1.3}
\caption{Performance of RCC with different settings.}
\label{table_example}
\centering
\scalebox{0.8}{
\begin{tabular}{cc|cccc}
\toprule
 ${\boldsymbol f}$  &  ${\boldsymbol g}$ & \bfseries Cora & \bfseries Citeseer & \bfseries Fb & \bfseries Weiz.\\
\midrule
logistic & mode & 0.787 & 0.770 &  0.745 & 0.802\\
{\bf logistic} & {\bf prop.} & 0.811 & {\bf 0.773} & {\bf0.748} & 0.806\\
logistic & sum & 0.806 & 0.765 & 0.748& {\bf 0.807}\\
\addlinespace
softmax & mode&  {\bf 0.818} & 0.766 & 0.747&0.805\\
softmax & prop. & 0.815 & 0.764 & 0.745& 0.807\\
softmax & sum & 0.817 & 0.766 & 0.743& 0.795\\
\bottomrule
\end{tabular}}
\label{tab:fg}
\end{table}

\paragraph{Comparing Relational Features and Classifiers}

\begin{figure*}[htbp]
\centering
\begin{minipage}[]{0.13\textwidth}
\begin{center}
\tiny Original Images\\
\includegraphics[trim={2.5cm 1.8cm 2.5cm 0.6cm},clip,width=1\textwidth]{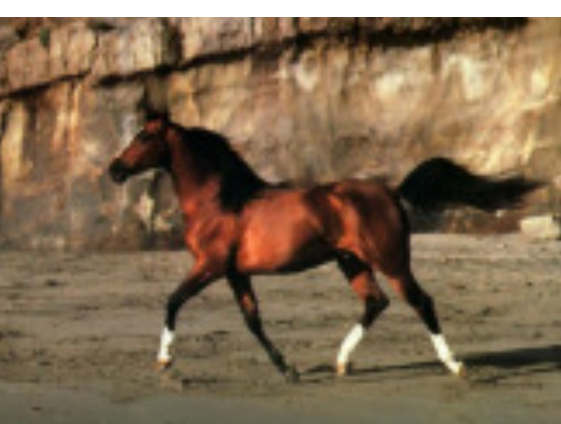}\\
\includegraphics[trim={2.5cm 1.8cm 2.5cm 0.6cm},clip,width=1\textwidth]{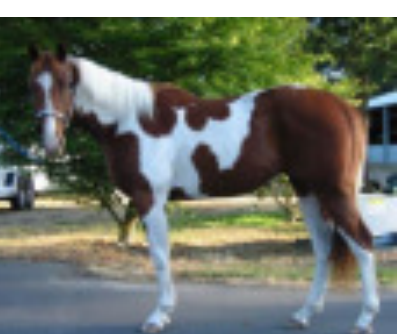}\\
\includegraphics[trim={2.5cm 1.8cm 2.5cm 0.6cm},clip,width=1\textwidth]{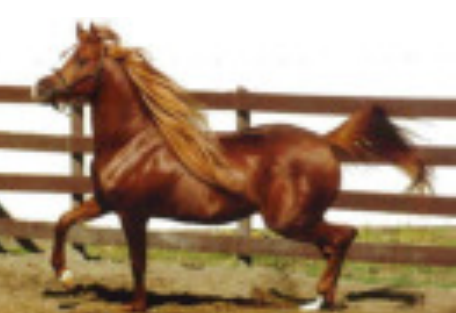} 
\end{center}
\end{minipage}
\begin{minipage}[]{0.13\textwidth}
\begin{center}
\tiny Noisy images\\
\includegraphics[trim={2.5cm 1.8cm 2.5cm 0.6cm},clip,width=1\textwidth]{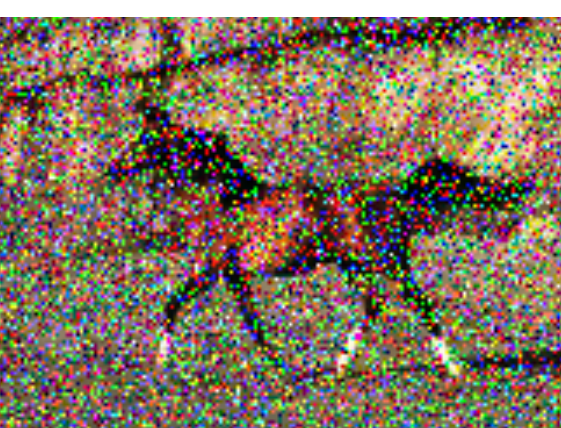}\\
\includegraphics[trim={2.5cm 1.8cm 2.5cm 0.6cm},clip,width=1\textwidth]{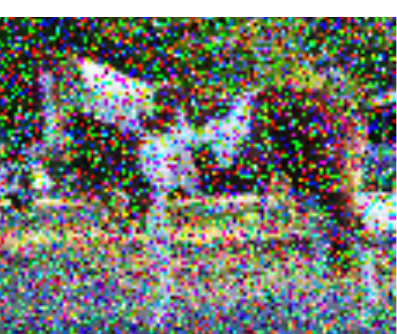}\\
\includegraphics[trim={2.5cm 1.8cm 2.5cm 0.6cm},clip,width=1\textwidth]{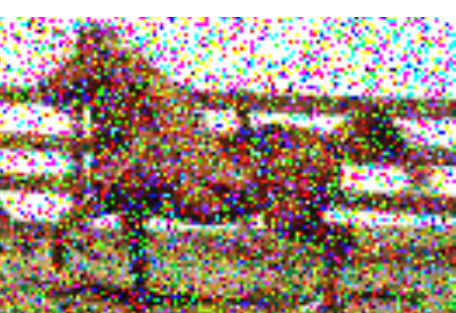} 
\end{center}
\end{minipage}
\begin{minipage}[]{0.13\textwidth}
\begin{center}
\tiny True labels\\
\includegraphics[trim={2.5cm 1.8cm 2.5cm 0.6cm},clip,width=1\textwidth]{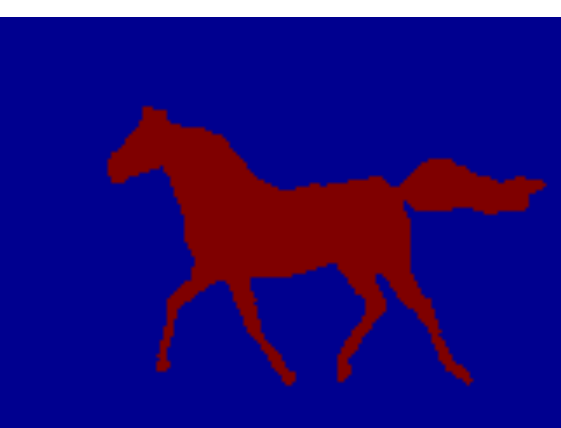}\\
\includegraphics[trim={2.5cm 1.8cm 2.5cm 0.6cm},clip,width=1\textwidth]{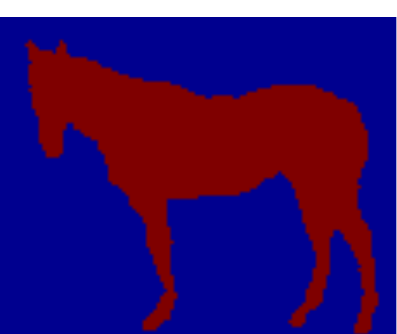}\\
\includegraphics[trim={2.5cm 1.8cm 2.5cm 0.6cm},clip,width=1\textwidth]{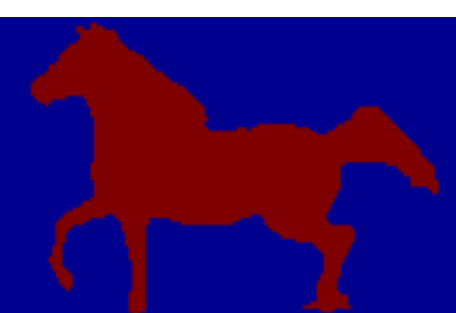} 
\end{center}
\end{minipage}
\begin{minipage}[]{0.13\textwidth}
\begin{center}
\tiny ICA\\
\includegraphics[trim={2.5cm 1.8cm 2.5cm 0.6cm},clip,width=1\textwidth]{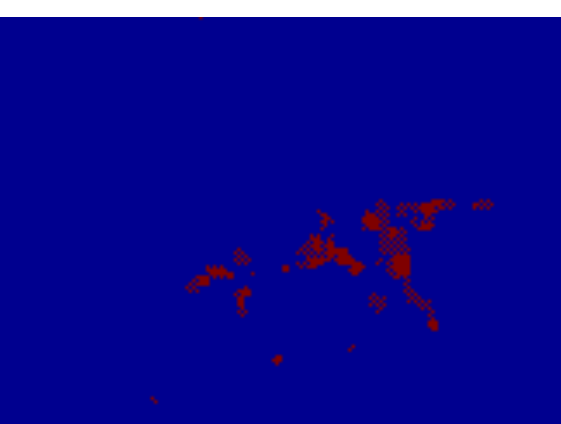}\\
\includegraphics[trim={2.5cm 1.8cm 2.5cm 0.6cm},clip,width=1\textwidth]{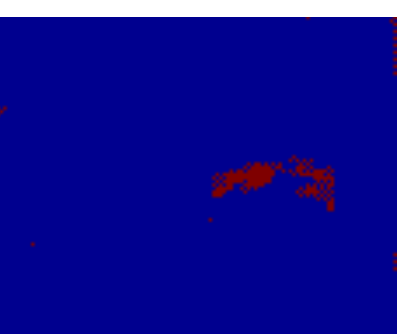}\\
\includegraphics[trim={2.5cm 1.8cm 2.5cm 0.6cm},clip,width=1\textwidth]{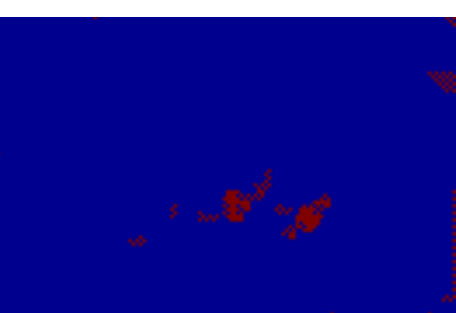} 
\end{center}
\end{minipage}
\begin{minipage}[]{0.13\textwidth}
\begin{center}
\tiny GS\\
\includegraphics[trim={2.5cm 1.8cm 2.5cm 0.6cm},clip,width=1\textwidth]{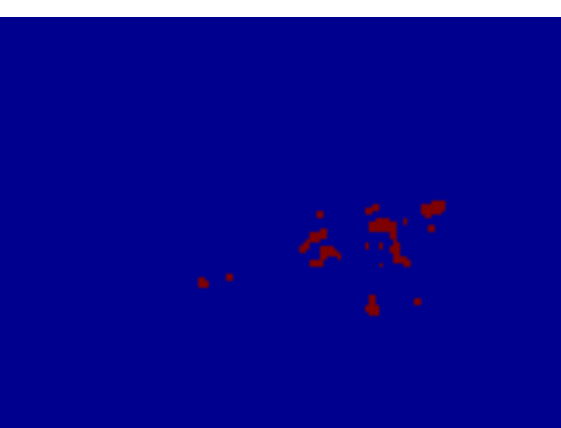}\\
\includegraphics[trim={2.5cm 1.8cm 2.5cm 0.6cm},clip,width=1\textwidth]{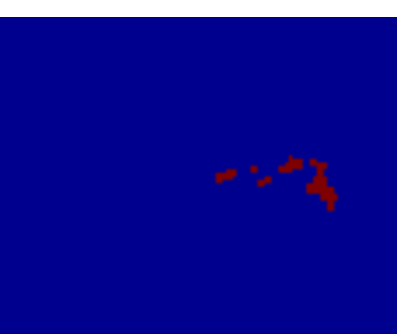}\\
\includegraphics[trim={2.5cm 1.8cm 2.5cm 0.6cm},clip,width=1\textwidth]{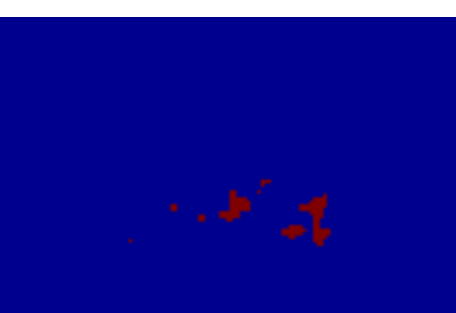} 
\end{center}
\end{minipage}
\begin{minipage}[]{0.13\textwidth}
\begin{center}
\tiny RCC\\
\includegraphics[trim={2.5cm 1.8cm 2.5cm 0.6cm},clip,width=1\textwidth]{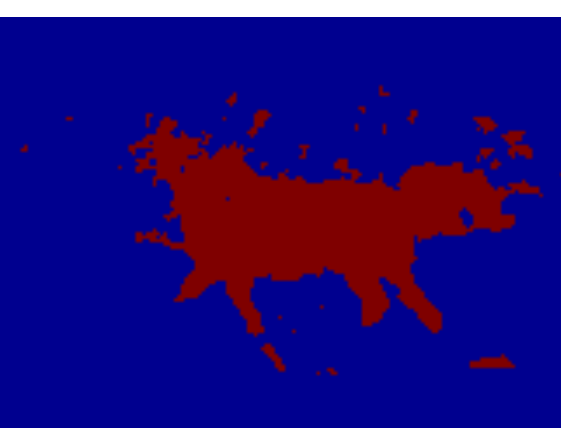}\\
\includegraphics[trim={2.5cm 1.8cm 2.5cm 0.6cm},clip,width=1\textwidth]{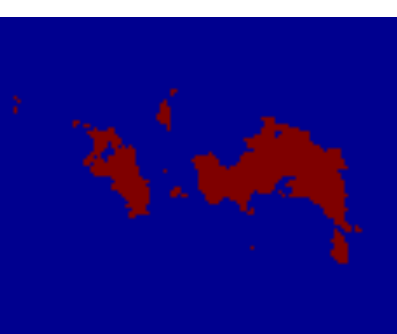}\\
\includegraphics[trim={2.5cm 1.8cm 2.5cm 0.6cm},clip,width=1\textwidth]{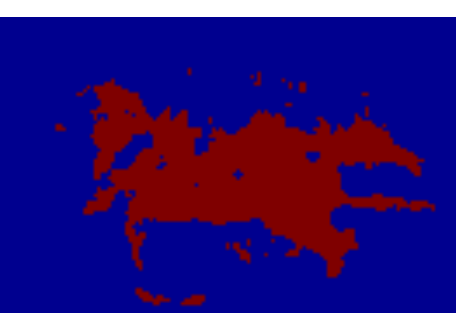} 
\end{center}
\end{minipage}
\begin{minipage}[]{0.13\textwidth}
\begin{center}
\tiny Local classifier\\
\includegraphics[trim={2.5cm 1.8cm 2.5cm 0.6cm},clip,width=1\textwidth]{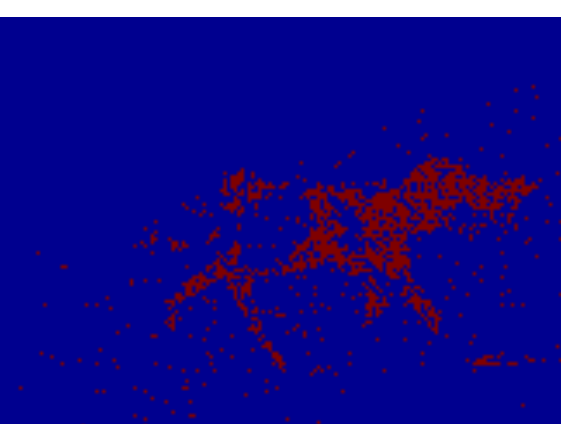}\\
\includegraphics[trim={2.5cm 1.8cm 2.5cm 0.6cm},clip,width=1\textwidth]{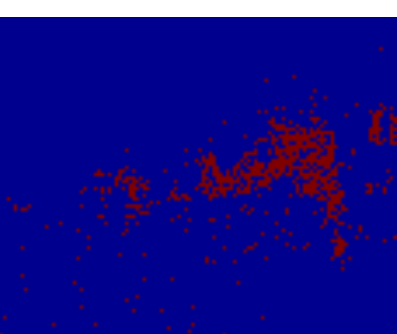}\\
\includegraphics[trim={2.5cm 1.8cm 2.5cm 0.6cm},clip,width=1\textwidth]{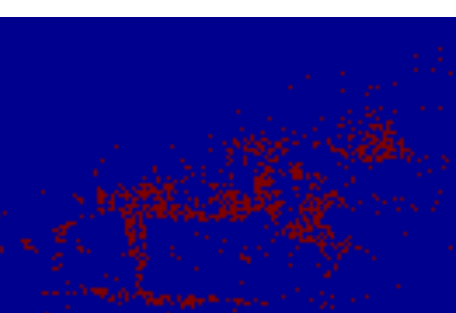} 
\end{center}
\end{minipage}
 \caption{Example segmentations using RCC, ICA, GS and local classifiers on Weizmann data. Because of the noisy local classifier, ICA and GS cascade errors into unreasonable segmentations, while RCC is more robust.}
  \label{fig:horses}
\end{figure*}

As we discussed before, RCC can be easily plugged in many local classifiers and it can also use many aggregation operators to include different relational features. To prove this, we run experiments with different classifier and aggregation operator for each of the four dataset: Cora, Citeseer, Facebook and Weizmann's dataset to compare the performance of different settings with RCC. For the local classifier, we choose from logistic sigmoid function and tempered softmax where we set ${\tau}$ to 0.5. For aggregation operators, we choose from mode, sum and proportion. For these experiments, we compare the average accuracy over 20 splits. The results in \Cref{tab:fg} show that for all these four datasets, there are not much differences between these six experiments. So for the following experiments, we use logistic sigmoid function and proportion.

\paragraph{Comparison of Prediction Accuracies}

The problem RCC aims to solve is the discrepancy between training ICA with relational features based on the true labels and the fact that at prediction time, ICA uses estimated labels. To illustrate this discrepancy, we run experiments that consider situations where local classification becomes more and more difficult. For the citation and network data, we generate versions of the data set where different fractions of the features are removed, in the range $[0.0, 0.9]$. For image data, we add varying amounts of salt-and-pepper noise. In effect, the experiments are run on versions of the data sets where prediction is harder, and more relevantly, where the assumption that the predicted labels are exactly the true labels becomes more and more incorrect. If RCC's training procedure is truly more robust to this scenario, we expect its improvement over ICA to become more pronounced as (local) prediction becomes more difficult.
The results for these experiments are shown in \Cref{fig:acc} where we plot the average accuracies or F-measures for the best-scoring regularization parameters over 20 splits of network data and 2 splits of image data. We compared RCC, ICA, GS, and the local classifier, which makes predictions only based on local features. The horizontal axis represents the amount of noise, and the vertical axis represents the training or testing accuracies achieved by these algorithms.

The results for all data sets suggest that RCC is more robust to weak local signal than ICA, GS, and local classifier. For the Cora and CiteSeer data, as the fraction of deleted features increases, the training accuracies of RCC stays stable until over $80\%$ of local features have been deleted. The training accuracies of ICA, GS, and the local classifier drop earlier and much faster. When $90\%$ of the features are deleted, the training accuracies of ICA, GS, and the local classifier drop to $0.5$, however the accuracies of RCC still remains around $0.9$, showing that RCC is able to train models to fully utilize the relational structure. The RCC test accuracies also show better performance than the other three methods as the number of local features reduces. Especially when over $60\%$ local features are deleted, the local predictors become less reliable which causes ICA's accuracy to significantly worsen, and RCC is able to withstand the lack of attribute information. The Facebook results follow similar trends, where the training and test accuracies are always better than the ICA and local classifier. The differences in the testing accuracy between RCC and ICA are statistically significant for all fractions of deleted features on the Cora tests, for 0.7 and 0.9 for the CiteSeer tests, and for all fractions 0.2 and higher on the Facebook tests. 

One interesting effect not often reported in other research is the tendency for ICA trained using the true labels to produce predictors that perform \emph{worse} than the local classifier, even on training data. This effect is exactly because of the discrepancy between the training regime and the actual prediction algorithm RCC aims to correct. For example, in all three of our data sets, there are settings, especially when the local classifier is noisy, that ICA has worse training accuracy than the local classifier. This effect is especially apparent in the image data, where the faulty assumption made by ICA causes it to consistently cascade local-classification errors, eventually leading ICA, GS, and the local predictor to predict that all pixels are background. RCC avoids this over-reliance on relational features, yet learns to incorporate relational features enough to improve upon the noisy local predictor. \Cref{fig:horses} contains example images and the predicted segmentations using the various learning algorithms. 

\section{Conclusion and Discussion}
\label{sec:discussion}

We presented recurrent collective classification, a variant of the iterative classification algorithm that uses differentiable operations, enabling back-propagation of error gradients to directly optimize the model parameters. The concept of collective classification has long been understood to be a principled approach to classifying nodes in networks, but in practice, it often suffers from making only small improvements, or not improving at all. One cause for this could be the faulty training procedure that we correct in this paper. Our experiments demonstrate dramatic improvements in training accuracy, which translate to significant, but less dramatic improvements in testing performance. RCC is a key step toward fully realizing the power of collective classification. Thus, an important aspect to consider to further improve the effectiveness of collective classifiers is the generalization behavior of collective models. One future direction of research is exploring how a more direct training loss-minimization interacts with known generalization analyses, perhaps leading to further algorithm improvements. Another future direction we are exploring is how to apply similar approaches of direct loss minimization in transductive settings and how to expand the flexibility of the RCC framework to incorporate other variants of ICA.

\bibliographystyle{plainnat}
\bibliography{fan-aistats17}

\end{document}